\pdfoutput=1

\documentclass[11pt]{article}

\usepackage{acl}
\usepackage{xspace}

\def\AC#1{{\color{magenta}AC: \it #1}}

\newcommand{\ism}[1]{\textit{ism~}\xspace}
\newcommand{\isms}[1]{\textit{isms}\xspace}

\usepackage{times}
\usepackage{latexsym}

\usepackage[T1]{fontenc}

\usepackage[utf8]{inputenc}

\usepackage{microtype}
 \usepackage{comment}

\usepackage{inconsolata}

\usepackage[compact]{titlesec}
\usepackage{url}
\usepackage{cleveref}

\usepackage[suppress]{color-edits}
\addauthor{ZT}{olive}
\addauthor{GA}{pink}

\newenvironment{myquote}[1]%
  {\list{}{\leftmargin=#1\rightmargin=#1}\item[]}%
  {\endlist}

\usepackage{todonotes}
\makeatletter
\newcommand*\iftodonotes{\if@todonotes@disabled\expandafter\@secondoftwo\else\expandafter\@firstoftwo\fi}  

\newcommand{\note}[4][]{\todo[author=#2,color=#3,size=\scriptsize,caption={},#1]{#4}} 

\newcommand{\zee}[2][]{\note[#1]{Z}{olive!40}{#2}}

\newcommand{\emo}[1]{\raise1.0ex\hbox{{\normalfont \textrm{\scriptsize #1}}}}

\title{Subjective \textit{Isms}? On the Danger of Conflating Hate and Offence\\ in Abusive Language Detection}

\author{Amanda Cercas Curry\thanks{\enspace Equal contribution.}\\
  MilaNLP \\
  Bocconi University \\
  \texttt{amanda.cercas@unibocconi.it} \\\And
  Gavin Abercrombie$^{*}$ \\
  School of Mathematical \\and Computer Sciences \\
  Heriot-Watt University \\
  \texttt{g.abercrombie@hw.ac.uk} \\\And 
  Zeerak Talat$^{*}$ \\
  Mohamed Bin Zayed \\ University of Artificial Intelligence \\
  \texttt{z@zeerak.org} \\}

\begin{document}
\maketitle

\begin{abstract}
Natural language processing 
research
has 
begun to embrace the notion of
annotator
\emph{subjectivity},
motivated by 
variations in labelling.
This approach 
understands each annotator's view as valid, which can be highly suitable for tasks that embed subjectivity, e.g., sentiment analysis.
However, 
this construction
may be inappropriate 
for tasks such as hate speech detection,
as it affords equal validity to all positions on
e.g., sexism or racism.
We argue that the conflation of hate and offence
can invalidate
findings on hate speech,
and 
call for future work to be situated in theory, disentangling
hate
from its orthogonal concept, 
offence.

\end{abstract}

\section{Introduction}



Recently,
natural language processing (NLP) 
researchers have
dedicated \ZTedit{significant} efforts towards 
tasks 
under the umbrella of online abuse
detection. 
\ZTdelete{\GAedit{---a phenomenon that encompasses \isms.}\ZTdelete{, including different \emph{isms}.}}
For example, racism~\cite[e.g.][]{talat-2016-racist,talat-hovy-2016-hateful}, sexism and misogyny~\cite[e.g.][]{jiang-etal-2022-swsr,zeinert-etal-2021-annotating}, xenophobia~\cite[e.g.][]{Ross_Measuring_2016}, homophobia~\cite{DiasOliva_Fighting_2021}, and transphobia~\cite[e.g.][]{chakravarthi-etal-2022-overview} have been all been proposed as 
suitable for automated identification using NLP methods.
Collectively these can be referred to as \isms~.
We understand \emph{isms} as 
prejudices, stereotyping, or discrimination on the basis on some personal characteristic.
For example, sexism is defined as prejudice, stereotyping, or discrimination, typically against women, on the basis of sex or gender~\cite{masequesmay-2008-sexism}.
\ZTdelete{Hate speech~\cite[e.g.][]{talat-2016-racist,talat-hovy-2016-hateful}, sexism and misogyny~\cite[e.g.][]{jiang-etal-2022-swsr,zeinert-etal-2021-annotating}, xenophobia~\cite[e.g.][]{Ross_Measuring_2016}, homophobia~\cite{DiasOliva_Fighting_2021}, and transphobia~\cite[e.g.][]{chakravarthi-etal-2022-overview} have been all been proposed as 
suitable for automated identification using NLP methods.}

\ZTedit{This line of research has been faced with high annotator disagreement~\cite[e.g.][]{leonardelli-etal-2021-agreeing}, and as a result has conceptualised this as an indication that the concepts themselves are subjective.}
\ZTedit{For example, \newcite{rottger-etal-2022-two} argue that 
labelling such phenomena is inherently subjective and can either be addressed as \emph{descriptive}, i.e., encouraging annotator subjectivity, or \emph{prescriptive}, i.e., discouraging it.
\ZTdelete{\ZTedit{Here, the descriptive paradigm is understood as `discouraging annotator subjectivity', while the prescriptive paradigm encourages it~\cite{rottger-etal-2022-two}. }\zee{Moved up from below.}}}
\ZTedit{By constructing abuse as individually subjective, 
social norms 
are
disregarded in favour of an approach that is blind to existing conditions of marginalisation.}
\ZTedit{This stands in contrast to early work in the field, which sought to tease apart the distinction between offensiveness and hate~\cite{Davidson_Warmsley_Macy_Weber_2017}, and sought frameworks to identify the particular vectors which indicated hate~\cite{talat-etal-2017-understanding,wright-etal-2017-vectors}.} 

\ZTedit{Discrimination is also an area subject to policy and regulatory debates.
Policy often distinguishes \emph{hate} from \emph{offence}.
For instance, in its definition of sexism, the European Institute for Gender Equality~\cite{eige} position sexism as the \emph{presence} rather than the offensiveness of a gendered stereotype\ZTdelete{, rather than the offensiveness of it}\ZTdelete{whether people find it offensive}:}
\ZTdelete{\ZTedit{Work within policy on gender equality, similarly, has sought to distinguish \emph{hate} from \emph{offence}:}}
\vspace{-0.22cm}
\begin{myquote}{0.5cm}
`Sexism is linked to beliefs around the fundamental nature of women and men and the roles they should play in society. Sexist assumptions about women and men, which manifest themselves as gender stereotypes, can rank one gender as superior to another.'
\end{myquote}
\vspace{-0.15cm}



\ZTedit{In this position paper, we consider 
such \emph{isms} and
how offence and hate\footnote{Hate speech `attacks or uses pejorative or discriminatory language with reference to a person or a group on the basis of who they are'~\citep{un-what}, including subtle stereotyping.} are orthogonal\footnote{We use `orthagonality' in the philosophical sense to refer to concepts that differ in scope, content, and purpose.} concepts that can be mutually informative, and argue that their conflation can delegitimise research artefacts and findings.
That is, we contend that 
the hatefulness of a statement
is invariant of a reader's position on whether it should be allowed within a particular public forum.
Consider for instance the use of gendered slurs: 
while inappropriate for a general audience (e.g., a public debate) 
they may be appropriate for others (e.g., academic work exploring the uses of expletives).}
\ZTdelete{That is, that whether a text is hateful or not is dependent on the position of the person reading it.}
In particular, we argue that \isms~ are culturally defined, whereas offence 
is a subjective experience.
Thus, we argue that it the presence of a stereotype that determines if a statement is hate speech, rather than 
individual
perceptions of its offensiveness.
\ZTedit{Understanding \isms~ as culturally defined, and offence as individually subjective allows us to distinguish any offence caused to a reader from whether a message contains hate speech.}
We therefore
call for 
approaches to
annotating online abuse that delineate the degree of offence caused
from the 
phenomenon
itself. 

\ZTdelete{In this paper, we consider 
how offensiveness and \emph{isms} are orthogonal concepts which can inform one another when teased apart, or delegitimise methods, models, and findings when conflated. \zee{This is what I mean is needed but we need to do this work earlier.}
We further call for research to address the conflation by providing a set of questions to help develop annotation guidelines that can tease apart experienced or perceived offence from the \ism~ under study.} 

\section{Understanding Subjectivity} 
Recent efforts in NLP have 
constructed annotation as
subjective, 
without attending to what other fields have understood 
this to mean.
\emph{Subjectivity} has been posed as the reason why `humans (e.g. 
annotators) [are] sensitive to sensory demands, cognitive fatigue, and external factors that affect judgements made at a particular place and point in time' \cite{alm2011subjective}.
Philosophy, however, sees \emph{subjectivity} as concerning people's differing perspectives, formed by factors such as cultural and individual experiences~\cite{Solomon2005-SOLS-2}.
This 
implies that the only valid knowledge is based on personal experiences, thereby negating the existence of objective or communal truths.
In contrast, \emph{relativism} proposes that criteria of judgement are relative to a culture or society~\cite{baghramian2004relativism}.
For instance, while 
humour 
may
be subjective, we can understand concepts such as beauty 
to be culturally defined.


\ZTedit{Hate speech detection, in particular, has often been argued to be a subjective task~\cite[e.g.][]{almanea-poesio-2022-armis,basile-2020-it}.}
\ZTedit{
Under this framing,
researchers 
collapse the label classes \emph{offensive} \emph{hate speech}~\cite[e.g.][]{leonardelli-etal-2021-agreeing}, thereby further conflating these concepts.
}
For instance, \newcite{akhtar2021opinions} posit that `judging whether a message contains hate speech is quite subjective, given the nature of the phenomenon'.
When categories of abuse are described as subjective, we understand that there is no ground truth, and wider cultural norms do not impact what constitutes hate.
Within the concept of \isms~, we argue that is the wrong approach and that these are culturally defined.
That is, we argue that, for a stereotype or norm, there \emph{is} a ground truth given by the cultural and temporal context a statement is made in.

\subsection{Stereotypes as Socially-defined Artefacts}

\emph{Isms} are a term given to various forms of marginalization and concepts such as racism, sexism, transphobia, etc.
Such \isms~ rely on tropes and stereotypes about a target group~\cite{manne2017down}.
They describe beliefs about the way a group is and how it ought to be~\cite{ellemers2018gender}. 
Although stereotypes are held by individuals, \ZTdelete{there are reasons to believe} they are formed collectively\ZTdelete{ by a group} \cite{Butler_Excitable_1989}. 
For example, stereotypes are observable: we can catalogue the content of gender stereotypes within a culture~\cite{prentice2002women}, suggesting these are not \ZTdelete{in fact (entirely) down to the}\ZTedit{solely} individual \ZTdelete{and}\ZTedit{but} instead exist in the `collective brain'. 

\ZTedit{\newcite{haslam1997group} argue that stereotypes emerge when individuals are acting in terms of a common social identity.
Although the belief that stereotypes are simply an inferior representation of an unfamiliar group may be alluring, they serve to represent group-based realities: they represent (and accentuate) perceived differences between then in- and out-group~\cite{haslam1997group}.
}
Through the lens of self-categorisation theory, \newcite{haslam1997group} argue that stereotypes are a social force--they reassure individuals of their belonging to a group `by: (1) enhancing perceived in-group homogeneity; (2) providing associated expectations of mutual agreement; and (3) producing pressure to actively reach consensus through mutual influence'. 
Uniformity of belief is thus the very essence of \ZTedit{a stereotype}\ZTdelete{stereotype content}. 
\ZTedit{The shared nature of stereotypes is what causes their severity, a single individual holding and acting on discriminatory beliefs is less consequential than a group holding and acting on the same beliefs}.
\ZTedit{However, because stereotypes are collective, they are also fuzzy; while individuals in the in-group are at least aware of stereotypes, they do not necessarily believe in them.
This is in part why the degree of offence to \isms~ may vary.
}
\ZTdelete{This \emph{shared} nature is the precise danger of stereotypes---we respond to stigmatised groups in similar ways. 
One individual hating short people sucks but it's not that big of a deal, a whole society (particularly the more powerful group) systematically and consensually discriminating against short people becomes injustice.
All this is not to say that \emph{all} individuals within a group hold the same beliefs and stereotypes. 
Because stereotypes are collective, they are also fuzzy. 
However, individuals within an in-group are at least \emph{aware} of the stereotypes that exist about other groups.} 
Group memberships and social relations play a key role in shaping cognition, leading to the application 
and salience of stereotypes to be context-dependent but consensual at the group level nonetheless.


\subsection{Acceptability as a Social Norm}


\ZTdelete{In political theory, the Overton window\footnote{See also Hallin's spheres \cite{}.} describes a spectrum of acceptable policies and discourse. }
Generally speaking, some \emph{isms} are less socially acceptable nowadays than they were a century ago due to the social justice movements of the last century.  
Such movements have, in some countries, resulted in an increased public awareness 
of the harms caused by stereotypes, making support for some of them less socially acceptable.
That is, the Overton Window, a political theory that describes the spectrum of acceptable policies and discourse, has shifted to make it less socially acceptable to hold particular stereotypical beliefs.
The result of such a shift is that people do not wish to label statements they agree with as an \ism, lest they be labelled as \emph{*ists} themselves.
For instance, homophobia has become less tolerated in many countries, and individuals do not want their statements, or them, to be labelled as homophobic.
Yet while being labelled as homophobic is perceived as undesirable, this does not mean that homophobic comments are not made, and policies not pursued. 
For example, in the United States of America, the American Civil Liberties Union has currently flagged more than 500 legal bills as anti-LGBTQ~\cite{ACLU_Mapping_2023}.
Thus, despite forward progress on some forms of discrimination and isms~\cite{gender_snapshot_2023,pew_racism_2023}, there are still socially acceptable \emph{isms} that come in two general flavours: the benevolent \emph{isms} and the scientific \emph{isms}. 

\ZTdelete{We don't want to label something as an \emph{ism} if we might find ourselves agreeing with it, lest we be found out as \emph{*ists} ourselves. 
However, there are still socially acceptable \emph{isms} that come in two general flavours: the benevolent \emph{isms} and the scientific \emph{isms}. }

\paragraph{The Benevolent *\emph{Ism}} 
Some stereotypes may be seen as `positive' and therefore not recognised by some as hateful.
The existence of `benevolent' stereotypes~\cite{jha-mamidi-2017-compliment}, \ZTdelete{also known as} such as `neosexism'~\cite{tougas1995neosexism}---those without clear negative connotations---means that annotators may be unlikely to recognise them as harmful. 
For example, the seemingly positive stereotype in Western nations that Asians are successful, high-achievers leads to their vilification (for being \textit{too} high-achieving) and the perception that they lack interpersonal skills~\cite{Wong_Model_2006}. 
These stereotypes may also cause indirect harm to the individuals who may feel they are not living up to what is expected from them~\cite{haslam1997group}.
We might be tempted to only oppose or target stereotypes that imply or directly state that a certain group is inferior, however this approach would leave many of the issues of stereotyping unaddressed.
For example, not addressing claims such as `women need to be protected' or that `women's bodies are more aesthetically pleasing' suggests that the perception of women as inferior, or inherently sexualised, should remain acceptable. 

\paragraph{The Scientific *\emph{Ism}} 
This \emph{ism} uses evolutionary biology as evidence for stereotypes.\ZTdelete{a stereotyping crutch.}
In this case, different groups are proposed as differing on the basis of \emph{natural} differences, such as physiology.
One such example is the idea that women are naturally more nurturing than men due to imaginations of gender roles of the past.
However, investigations of hunter-gatherer societies indicate that this idea may not be an accurate reflection of past societies and social evolution~\cite{Hewlett_Fathers_2010}.
The idea of evolutionary psychology as evidence stems from Social Darwinism~\cite{miller2011mating}, which argues that one cannot accuse nature of being \textit{-ist}, and therefore any generalisation based on biology cannot be labelled as such.
Such pseudo-scientific \isms~ are commonly used as a rationalisation for the `objective' differences between dominant and marginalised groups (e.g. \newcite{browne2006sex}).

\ZTdelete{Another contributor to the complication of the identification of hate speech is the notion that some things are not stereotypical, but rather \emph{natural} differences between groups, such as the idea that women are naturally more nurturing than men because of the supposed gender roles of our ancestors. 
This idea, stemming from Social Darwinism, purports that one can hardly say Nature is sexist, therefore any generalisations based on biology can be forgiven. 
These pseudo-scientific \isms~are less common outside of sexism because they have often been denounced during the 21st century as white supremacist nonsense, but they were also commonly used as rational, objective differences to other marginalised groups. \AC{Actually fact-check this? Just my impression}. }

\subsection{Separating \emph{Isms} and Offensiveness}
\ZTedit{
So far, we have established that \isms~ are rooted in socio-cultural contexts, and, while not necessarily 
factual or objective, 
exist as normative
and 
therefore stable concepts, given their socio-cultural and temporal situations.
As
norms, 
\isms~ can 
cause 
harms to members of targeted groups, 
present barriers to harmonious community relations, or pose threats to law and order~\cite{barendt2019harm}.
}

Offensiveness can 
be understood as moral outrage or disgust~\cite{sneddon2020offense}.
\ZTedit{As 
\isms~ can be harmful, it is tempting to suggest that they should always be constructed as offensive. 
However, 
this
would not afford the 
high levels of disagreement
often
observed in their annotation. 
Such disagreement can be accounted for by considering the degree of offence taken as subjective.
That is, the degree of offence is knowable only by each annotator.
}
According to \newcite{sneddon2020offense}, we tend to give claims of offensiveness more credence than they deserve.
People 
get
more offended about topics that particularly matter to them, and these are impacted by one's identity: A citizen of the USA is more likely to be offended by the burning of their national flag than a European.
That is to say, when we are offended, we take the object of offence as a personal affront. This has material consequences when it comes to modelling \isms~ as offensive.

%
%
%
%


\ZTdelete{Recently in natural language processing research, there has been a movement away from aggregating the multiple labels per item collected from different annotators to a single  majority judgement or `gold label'.
Instead, researchers are increasingly concerned with collecting, releasing, and modelling the different \emph{perspectives} of individuals and groups of annotators, and capturing disagreements between them e.g. \cite{nlperspectives-2022-perspectivist,leonardellli-etal-2023-semeval,plank-2022-problem}.} 

\ZTdelete{While this has often been motivated by a desire not to erase minority voices within the annotator pool \cite{nlperspectives-2022-perspectivist,blodgett-2021-socio}, is is also often claimed that the disagreements are the result of the annotation tasks being subjective.
This notion of subjectivity has been claimed for a wide range of tasks, including linguistic tasks such as anaphora resolution, part-of-speech tagging and natural language inference/textual entailment \cite{abercrombie2023consistency}.
However, it is more often applied to social tasks such as sentiment analysis and identification of \emph{isms} such as detection of hate speech and related phenomena.}

\ZTdelete{For the latter, some researchers have described hate speech as being inherently subjective.
For example, Akhtar et al. \cite{akhtar2021opinions} claim that `judging whether a message contains hate speech is quite subjective, given the nature of the phenomenon', Almanea and Poesio \cite{almanea-poesio-2022-armis} consider annotator disagreements on labelling of sexism and misogyny to be subjective, Basile et al. \cite{basile-2020-it} provide hate speech detection as an example of a `highly subjective task', and \cite{sandri-etal-2023-dont} investigate reasons for what they consider to be subjective disagreement in hate speech annotation.
}

\ZTdelete{Beyond such descriptions, the creation of datasets for these tasks in some cases constructs the target phenomena as subjective.
For example, Leonardelli et al. \cite{leonardelli-etal-2021-agreeing} blur the distinction between hate speech and offensiveness by merging the \emph{offensive} and \emph{hateful} classes from different sources as one target phenomenon which they describe as `subjective'.
In actual fact, they attempt to cast offensiveness detection as an objective task, by instructing annotators to judge the examples impersonally.\footnote{An excerpt from the annotator guidelines reads: `Try to judge the offensiveness of the tweets independently from your opinion but solely based on the abusive content that you may find.'} 
And R{\"o}ttger et al. \cite{rottger-etal-2022-two} provide hate speech annotation as the example of a task that can be performed descriptively---that is, with little or no guidance, and according to individual interpretation---due to its supposed subjectivity.}



\section{Annotator Competency}
\ZTedit{Dataset labelling in NLP is typically performed by annotators recruited either as crowd-sourced workers~\cite[e.g.][]{abercrombie-etal-2023-temporal,basile-etal-2019-semeval,fersini2018overview}, academics or students available to the researchers~\cite[e.g.][]{cercas-curry-etal-2021-convabuse,fanton-etal-2021-human,jiang-etal-2022-swsr}, or people deemed to hold expertise in the phenomena~\cite[e.g.][]{talat-2016-racist,vidgen-etal-2021-learning,zeinert-etal-2021-annotating}.
However, the Standpoint Theory~\cite{harding-1991-whose} argues that annotators, can largely only be competent within their own lived experiences, regardless of training.
Without lived experience, annotators may not be able to gain a full understanding of the \ism~ under consideration.
For instance, \newcite{larimore-etal-2021-reconsidering} found that white annotators were far less competent in identifying anti-Black racism than Black annotators.
Guidelines and labelling taxonomies, no matter how thoroughly and carefully constructed are not capable of adjusting for a lifetime of lived experience.
It is not, therefore, inherent subjectivity within the task, but rather differences in annotator ability due to their personal standpoint that impact on annotators' ability to recognise whether hate speech or abuse is present.
Sometimes even if an individual does recognise the target phenomenon, they may choose to ignore it for political reasons~\cite{marable1995beyond}.
}
\ZTdelete{In NLP research, including abusive language and hate speech detection, dataset labelling is typically performed by annotators recruited in one of a number of ways: crowd-sourced workers e.g.~\cite{abercrombie-etal-2023-temporal,basile-etal-2019-semeval,fersini2018overview}, academics or students at the researchers' institutions or in their networks e.g.~\cite{cercas-curry-etal-2021-convabuse,fanton-etal-2021-human,jiang-etal-2022-swsr}, people deemed to have expertise in the target phenomenon, or people recruited and trained specifically for the task e.g.~\cite{vidgen-etal-2021-learning,zeinert-etal-2021-annotating}.
However, according to \emph{Standpoint Theory}~\cite{harding-1991-whose}, no matter who they are or how they are trained, to a large extent, annotators can only be competent within their own lived experiences.
For example, Larimore et al. \cite{larimore-etal-2021-reconsidering} found that white annotators were far less likely to recognise anti-Black racism than were Black annotators.
Similarly, Sap et al. \cite{sap-etal-2022-annotators} found that white annotators were far less able to recognise the benign nature of in-group use of African-American Vernacular English then their Black counterparts.}

\ZTdelete{However they may be recruited and trained, individual annotators do not necessarily 
have a full understanding of the \emph{ism} that they are labelling.
Annotator guidelines and labelling taxonomies, no matter how thoroughly and carefully constructed are not capable of adjusting for a lifetime of lived experience.
It is not, therefore, inherent subjectivity within the task, but rather differences in annotator ability due to their personal standpoint that impact on annotators' ability to recognise whether hate speech or abuse is present.
Sometimes even if an individual does recognise the target phenomenon, they may choose to ignore it for political reasons~\cite{marable1995beyond}.
}


\section{
Towards a New Formulation of 
\emph{Isms} as Cultural Formation of Societal Norms}

Given our understanding of \isms~ as culturally relative constructions and \emph{offence} as an individually subjective concept, we propose that \isms~ can best be understood as cultural formations of societal norms.
That is, \isms~ encode norms, which are inherently fuzzy at the border~\citep{Hall_Race_1997}.
When creating data for \isms~, researchers often work at the fuzzy borders of acceptability.
In operating at these borders, and developing computational methods to draw them, research delineates what is acceptable from that which is not. 
While such borders are inherently messy, through an understanding of determining acceptability as cultural norms, we can refocus our attention towards the question of how such norms and borders should be drawn.

For instance, \citet{Douglas_Purity_1978} argues that determining what is `dirt'
is a cultural process which strengthens communities and builds community cohesion.
That is, while encountering an offensive instance, i.e., an instance of sexism, can be destabilising to a community, the process with which the community makes a determination, and the determination itself, allows for the community to reify itself.
This is particularly important as we can come to understand that \isms~ are culturally defined objects, and identifying the borders of acceptability necessitates an ongoing negotiation with the communities in question~\cite{thylstrup-talat-2020-detecting}.
Within this formulation of \isms~, we can come to understand \isms~ as distinct from \emph{offence}. 
Thus, this formulation of \isms~ provides space for both a cultural understanding of \isms~ whilst making space for \emph{offence} as an individual and subjective notion.

\section{Recommendations}

We have argued that 
conflation of \isms* and offence
stems from annotation \textbf{task construction}. 
We 
recommend that 
schema 
be designed to
carefully delineate these concepts,
by e.g., creating 
distinct categories, and labelling them separately.

As 
guidelines cannot meaningfully offset gaps between annotators and any missing lived experience required to identify \isms~, we recommend that \textbf{annotator recruitment} 
target people with relevant profiles to label the data in question.
Where 
this is not possible,
schema
should
allow annotators the option of indicating where they do not have the necessary lived experience to label specific items.



\section{Conclusion: Implications for NLP} 

If, as we propose, 
identifiying \emph{isms} 
is not subjective, we must conclude that annotator differences are irrelevant at the individual level for such tasks.
Rather, they are symptoms of disagreement on the degree to which \emph{isms} offend individual annotators.


At the group level, we must take care not to treat conflicting responses equally.
If 
a minority 
with 
the necessary lived experience (e.g. to recognise misogyny) 
disagree with the majority who don't, 
that matters.
For example, \citet{gordon-etal-2022-jury} attempt to pick out the `correct' minority perspectives from the wider pool of annotators for each instance,
and
\citet{fleisig-etal-2023-when} specifically assume that the majority of annotators are likely 
`wrong', 
i.e.,
they will not recognise the target phenomenon.

Construction of the desired classification schema based on societal norms comes with its challenges.
While prescriptivist annotation based on agreed societal norms may be desired, it can be difficult or even impossible to implement comprehensively in practice.
One reason for this is that it is probably not possible to recruit annotators with the correct standpoint or competencies to recognise every instance---or indeed to know what those characteristics might be. 
Another is the nature of building classification schema.
While a clearly defined, unambiguous, comprehensive and static \emph{Aristotelian} classification scheme may be desired 
rather than
\emph{prototypical} classification, 
it can be hard or even impossible to implement, and people generally resort to the latter~\cite[p. 61-62]{bowker-star-2000-classification}.

Despite this, 
we believe that it is vital that 
\emph{isms} like misogyny and other hate 
and abuse not be constructed as individually subjective, but rather as culturally formed societal norms.
While there may be much to gain from examining the responses of individual annotators to these tasks, NLP researchers should be careful not to conflate individual differences with inherent subjectivity of tasks.

\section*{Limitations}
We have presented a position on the modelling of hate speech in NLP backed by existing literature in philosophy, gender studies, and critical race theory. While we have made actionable recommendations for NLP researchers working on hate speech and related phenomena, schema definition and annotator recruitment to exactly capture a phenomenon are known to be challenging. We encourage researchers to follow best practices and involve interdisciplinary researchers given the nature of the task.\looseness=-1

\section*{Ethical Considerations}
This paper presents a re-framing of tasks related to hate speech and abusive language detection. 
In this new frame, we delineate between that which causes offence at an individual level and that which is hate, 
defined at a societal level with regard to concepts such as sexism, racism, and so forth, collectively referred to as \textit{isms}.
From this understanding of \isms~, it becomes clear that current practices reinforce social norms of desirability and respectability.
The implications of disentangling offence from \isms~, is then to disentangle individual desirability from our understanding and modelling of \isms~.
Consequently, our framing 
makes space for marginalised communities to name the discrimination that they are subject to, without also making determinations on whether discriminative messages should be moderated for \emph{all} potential viewers.
This
affords space for marginalised communities, in particular, to 
call out
the discrimination that they are subject to, 
regardless of whether others recognise that discrimination.
Furthermore, by disentangling offence from \isms~, public policy analysis and decisions on what should be regulated and what should be subject to individual preference can disregard whether content causes offence, and instead pay attention to whether the content constitutes a discriminatory statement on its own merits. 
Data and models that arise from disentangling offence from \isms~ thus afford individuality in terms of what causes offence to an individual, and therefore what they would wish to (not) be exposed to, without making inference as to whether that content constitutes an \textit{ism}.
Further, our framing of \isms~ removes sovereignty to individually define and operationalise \isms~.
Instead, we follow \citet{Butler_Excitable_1989} in their understanding that \isms~ arise from the socio-cultural citations of past events, i.e., from the norms that are established and reused in a given society over time.
Thus, establishing what constitutes an \ism~ is a task that must be conducted by examining the social and political conditions in a given society and is 
liable
to change with society.\looseness=-1

\section*{Acknowledgements}
Amanda Cercas Curry was supported by the European Research Council (ERC) under the European Union’s Horizon 2020 research and innovation program (grant agreement No.\ 949944, INTEGRATOR). She is a member of the MilaNLP group and the Data and Marketing Insights Unit of the Bocconi Institute for Data Science and Analysis (BIDSA).
Gavin Abercrombie were supported by the EPSRC project `Equally Safe Online' (EP/W025493/1). Gavin Abercrombie, Tanvi Dinkar and Verena Rieser were supported by the EPSRC project `Gender Bias in Conversational AI' (EP/T023767/1).








\bibliography{anthology,custom}




\end{document}